# Exploring Explainable AI in the Financial Sector: Perspectives of Banks and Supervisory Authorities

Ouren Kuiper[1][0000-0002-5033-6173], Martin van den Berg[1][0000-0003-3974-7374], Joost van der Burgt[2], and Stefan Leijnen[1]

[1] HU University of Applied Sciences Utrecht, Utrecht, Netherlands
{ouren.kuiper, martin.m.vandenberg, stefan.leijnen}@hu.nl
[2] De Nederlandsche Bank, Amsterdam, Netherlands

**Abstract.** Explainable artificial intelligence (xAI) is seen as a solution to making AI systems less of a "black box". It is essential to ensure transparency, fairness, and accountability – which are especially paramount in the financial sector. The aim of this study was a preliminary investigation of the perspectives of supervisory authorities and regulated entities regarding the application of xAI in the financial sector. Three use cases (consumer credit, credit risk, and anti-money laundering) were examined using semi-structured interviews at three banks and two supervisory authorities in the Netherlands. We found that for the investigated use cases a disparity exists between supervisory authorities and banks regarding the desired scope of explainability of AI systems. We argue that the financial sector could benefit from clear differentiation between technical AI (model) explainability requirements and explainability requirements of the broader AI system in relation to applicable laws and regulations.

## 1 Introduction

In recent years increasingly powerful, but often also increasingly complex, machine learning methods have become available and are used to greater extent in commercial contexts [1,2]. Generally, this form of machine learning is referred to simply as "artificial intelligence" (AI). The increasing use of novel and hard-to-understand types of AI systems has sparked a discussion on the need for explainability of AI [3,4]. Especially for high-risk use cases there is a realization, both scientifically and societal, that AI needs to be explainable to be understood. For instance, the upcoming EU legislature on AI [5] will require demonstrable transparency for which explainable AI will be essential. In the financial sector comprehensive understanding of the use of AI systems is even more crucial: both stipulated by a wide range of laws and regulations and because trust in financial institutions is of high importance [6]. Simultaneously, expectations of new AI systems are high in the financial sector, while regulators need time to keep up with the speed of development [7]. Striking the right balance between performance and explainability can present a difficult dilemma for financial institutions.

The field of explainable AI (or 'xAI') studies how AI can be made explainable by making algorithms and their systems more transparent, often referred to as "opening the black box" [3]. An improved understanding of the working of these algorithms helps us to verify them, improve them, and implement them ethically. Most developments in xAI focus on either technical tools for model developers or approach explainability as a social or cognitive challenge [8,9]. Other authors have stated that making models explainable should be foregone instead of using inherently interpretable models [10]. Given the attention transparency and explainability receive as a requirement for ethical AI, it is no surprise that many reports on the responsible use of AI have stressed the need for xAI [11]. Notably, the number of empirical studies that provide practical insights into how xAI is actually used in practice is very limited [12] which we believe represents a hiatus in the current literature.

Financial institutions, both large and SME, have begun to use AI, for instance in delivering instant responses to credit applications, claim settlement, and transaction monitoring [24,25]. The World Economic Forum [16] notes that the opacity of AI systems poses a serious risk to the use of AI in the financial sector: lack of transparency can lead to loss of control by financial institutions and thereby damage consumer confidence and society. Given the crucial role of trust in the financial sector, explainability of the outcomes and functioning of AI systems is considered necessary [16]. Explainability is in fact one of the EU's key requirements for trustworthy AI [11]. With new EU AI legislation announced, explainability is expected to become even more important and necessary for some high-risk use cases such as consumer credit scoring [5].

Limited empirical descriptions on the challenges surrounding the application of xAI exist. In addition, only preliminary guidelines exist [17] on how to implement xAI, often based in theory and lacking empirical validation. In the future, a solid and practical framework could help organizations to better understand their obligations (regulatory and otherwise) regarding xAI and how to operationalize them. In the financial sector, such a framework could also help supervisory authorities to translate current regulations regarding transparency and the provision of information, to clear expectations regarding xAI to regulated entities. In lieu of such a framework, a starting point is to map what is currently expected of in terms of explainability of AI by banks and supervisory authorities.

The current exploratory study aims to identify what the differences are regarding the expectations of explainability of AI for supervisory authorities and regulated entities in the financial sector. Three use cases were examined in which AI is used at financial institutions in the Netherlands. Data were collected by means of semi-structured interviews with interviewees of both banks and supervisory authorities. This study is intended to add empirical data on how xAI is regarded and used in practice and as stepping stone towards a framework as described above. The main research question is: *What are the perspectives of supervisory authorities and regulated entities regarding the application of xAI in the financial sector?*

## 2      Theoretical background

Explainable AI (xAI), also referred to as interpretable or understandable AI, aims to solve the "black box" problem in AI [18,19]. A typical present-day AI system utilizes data (e.g., information on a person's financial situation) and produces an outcome (e.g., a risk of default indication). However, in such a system it is not always evident from the output how or why a certain outcome is reached based on the data. Especially when using more complex AI systems (e.g., using deep learning or random forest methods) the process from input to output is practically impossible to understand by humans even with full knowledge of the inner workings, weightings, and biases of the system. The term xAI encompasses a wide range of solutions that explain why or how an AI system arrives at outcomes or decisions [20]. One line of research focuses on technical tools to explore the relation between model input and output, such as SHAP [21] and LIME [22]. A critique on the xAI field expressed by various authors is that xAI is often not clearly defined and discussed without proper understanding of the surrounding concepts and the parties involved [19,23]. As such, the exact scope of xAI is not always well-defined, as sometimes the term is used to focus on technical solutions directly relating to the model, but sometimes the system context is also taken into account.

Transparency is one of the central concepts of xAI. Importantly, the term is used in two distinguishable contexts or manners in the literature, which we differentiate by using *model transparency* and *process transparency*. Model transparency is the property of a model to be understood by a human as it is, in terms of its general working or design. The opposite of "black-boxness" is model transparency [3,10]. This type of transparency is generally what model developers refer to and is highly related to the concept of interpretability [24,18]. Process transparency is transparency of the use and development of an AI system; it relates to openness and not concealing information for stakeholders [24]. This form of transparency is generally what the colloquial meaning of transparency refers to. However, it is also the type of transparency that is meant in some of the literature on responsible use of AI when talking about "transparency" [10,17].

Explainability means that an explanation of the operation and outcome of a system can be formulated in such a way that it can be sufficiently understood by the stakeholder [3]. The term "stakeholder" refers to the individual, party, or audience impacted by the functioning and/or outcomes of the AI system, requiring information in the form of an explanation. In a vacuum, i.e., without a stakeholder, an explanation cannot be said to do what is intended, namely making something understood by an individual [9]. We would argue that the core concept of explainable AI is *effectual* explanation. An effectual explanation is not only about providing the required information, but to do so in a manner that leads to stakeholder understanding [25], for instance by offering the right amount of detail or boundary conditions of a model [26]. In addition, explanations can be global or local [13,14,26]. That is, a global explanation reveals the inner workings of the entire AI system (potentially including a case at hand), a local explanation offers insight in a specific outcome.

We used the following definition of explainable AI in this study: "*Given a stakeholder, xAI is a set of capabilities that produces an explanation (in the form of details,* *reasons, or underlying causes) to make the functioning and/or results of an AI system sufficiently clear so that it is understandable to the stakeholder and addresses the stakeholder's concerns."* [15].

Various types of information that can be used as the basis for an explanation can be distinguished. A distinction that should be noted here is that of the of process-based versus outcome-based explanation [17]. A process-based explanation gives information on the governance of the AI system across its design and deployment; the explanation is about "the how". An outcome-based explanation gives information on what happened in the case of a particular decision; the explanation is about "the what". In addition, explanations can be said to be "global" (explaining the entire model) or "local" (explaining a specific outcome) [13,14,26]. Furthermore, xAI techniques to gain more information about the functioning of a model can be model-agnostic (and work on any model, e.g., SHAP [21]), or be model-specific.

As a basis for this study we established a list of types of information that can underpin an explanation (of an AI system) that are relevant in the financial sector. We based this list on literature on explainable AI (using snowball search and focusing on the most cited papers in the field) and adapted it to fit use cases in the financial sector ([9,13,14,17,26]) . It should be noted that we incorporated types of information that are related to process-based explanation (e.g. the process surrounding the AI system), and which might be omitted in some views of explainable AI, that are however relevant from a regulatory perspective on AI in finance.

- The reasons, details, or underlying causes of a particular outcome, both from a local and global perspective.
- The data and features used as input to determine a particular outcome, both from a local and global perspective.
- The data used to train and test the AI system.
- The performance and accuracy of the AI system.
- The principles, rules, and guidelines used to design and develop the AI system.
- The process that was used to design, develop, and test the AI system (considering aspects like compliance, fairness, privacy, performance, safety, and impact).
- The process of how feedback is processed.
- The process of how explainers are trained.
- The persons involved in design, development, and implementation of the AI system.
- The persons accountable for development and use of the AI system.

## 3 Research method

### 3.1 Use cases

To address our research question, we applied a qualitative research approach by means of a series of semi-structured interviews. Three types of use cases were examined. The two supervisory authorities took part in all three use cases, with each of the three banks partaking in two of the three use cases (due to constraints in availability of interviewees). The three use cases were: 1) consumer credit, 2) credit risk management, and 3) anti-money laundering. A brief outline of these use cases will now be given.

The use case on consumer credit considers a typical case for consumer credit and a mortgage lending case. Consumer credit is credit provided to a consumer, which can be used to purchase goods and services. Financial institutions that provide consumer credit in the Netherlands have the right and obligation to ensure that the borrower has the capacity to repay the loan. Credit risk management focusses on internal risk and/or capital requirement models (early warning systems and probability of defaults models) where AI systems can be used to improve or replace the currently used models. The use case on anti-money laundering (AML) concerned AI systems which are used to conduct suspicious activity monitoring and transaction monitoring.

### 3.2 Data collection

The organizations involved in this study are two supervisory authorities (SAs) and three banks in the Netherlands. For reasons of anonymity these will be referred to as "SA", or "first SA", "second SA", "first bank", etc. depending on which interview took place first. The three banks belong to the major banks in the Netherlands, each with more than one million clients, and can be characterized as financial incumbents [27]. Semi-structured interviews were conducted with employees of these five organizations regarding the three use cases. For all interviews, use case experts (i.e., individuals that worked primarily on the use case at hand) were present. These experts either had a technical expertise (those directly involved with the development of the AI system) and/or a more supervising/governing role (such as compliance & risk officers and model owners).

At each interview at least two interviewees of that organization were present, and at most four (if the complexity of the use case required more diverse expertise in the interviewees). Interviews took between 1 and 1.5 hours. In total 13 interviews took place, six with interviewees from supervisory authorities and seven with interviewees from banks (as one bank took part in an additional interview to fully cover all questions). In addition, the findings were refined in a plenary session in which at least one participant of all five organizations was present. As a starting point during the interviews, a list of questions was used to guide the discussion, but the conversation was permitted to develop naturally in the direction deemed most suitable by the interviewers and interviewees.

The interviews with the banks and supervisory authorities had a slightly different list of starting questions, as the SA interviewees did not have the same direct knowledge of a specific use case in contrast to the banks. The interviewees of the banks were asked questions about the following topics: the context of how AI is being used in the organization, the role of explainability in the AI development process, the workings of the use case at hand, the application of AI in the use case, the relevant stakeholders, and how the bank deals with explainability in this particular use case. Finally, the banks were asked what types of information that can serve as a basis for explanations (based on the list from section 2) are considered relevant for supervisory authorities.

For the supervisory authorities, the focus of the interviews was on the boundaries of what they would allow in terms of AI and what their expectations of explainability were for that use case. The interviewees were asked questions concerning: their perception of the use of AI and xAI, applicable legislation around the use case, and the requirements for explainability from a supervisory perspective. In addition, they were asked what types of information (based on the list from section 2) they consider relevant for their supervisory role for the use case at hand. The interviews with the two supervisory authorities were conducted with interviewees who were aware of the applicable prudential, integrity and conduct regulations relating to the use cases.

All interviews were conducted by two researchers of the HU University of Applied Sciences via Webex. During every interview, one of the researchers had the lead in asking questions while the other made notes used for later analysis. After the interviews, the interviewees verified the interview reports and supplemented information where needed.

### 3.3 Data analysis

Data analysis was conducted based on the interview reports. As a first step we analyzed the interview reports and created a list of the main findings and conclusions per interview. These findings and conclusions were verified and supplemented by the interviewees. As a next step, we analyzed all interview reports and developed an overview of the main conclusions. These conclusions were discussed in a plenary session with participants of the supervisory authorities and banks. The output of this session was used to refine the conclusions.

## 4 Results

First, we discuss the most notable results per use case. Second, we discuss the overall findings.

### 4.1 Consumer credit

The first bank provided a use case about mortgage lending (a type of consumer credit) in which an AI system was used to assess mortgages with traffic-light colors to support middle office employees. The AI system runs in parallel to other, more traditional systems in the mortgage approval and monitoring process (e.g. using business rules). The AI system uses a rather simple form of machine learning based on logistic regression and uses around 10 variables. It improves on a business rules system in that it uses historical data. Interestingly, relating to explainability the primary users of the AI system (the middle office employees) were by design not given detailed insight into the functioning and results of the AI system to prevent potential gaming of the system. Due to the relative interpretability of the model, explainability to other stakeholders was not considered to be a challenge beyond the previous systems.

The second bank also supplemented their traditional loan approval system for consumer credit with an AI system. The traditional system uses basic data, such as the data a client provides through the application process or data from credit bureaus. The new AI system is trained and continuously supplied with new transactional data. The combination of both models resulted in fewer defaults on loans. For this use-case, model developers were considered the most important stakeholders regarding explainability. It was stated that it would be possible from a technological point of view to explain the model to customers, although this requires a thorough understanding of which type of narratives would be comprehensible by different consumer groups. This might require an interactive process, which was indicated to present a challenging IT problem rather than a problem of getting the relevant information (and explanations) from the AI system.

One of the SAs monitors whether lenders (i.e. banks) comply with lending standards. The lending standards ("leennorm" in Dutch) follow straightforward rules limiting the amount that can be loaned depending on the financial situation of the lender and are the basis for valid loan approval. Regardless of what an AI system indicates, banks must (and do) conform to this lending standard in all cases. The interviewee of the SA indicated that this was the primary method by which the supervisory authority currently ensured a lending consumer was protected. An interesting point was raised that within the lending standards banks might use AI to find cases their traditional systems would not give a credit, but the AI determines as being profitable for the bank. However, this might not always be good for the consumer. Widespread adoption of AI models might thus require reevaluation of the lending standards.

In summary, for consumer credit, banks reported they use AI in conjunction to traditional ("business rules") systems. As a result of the lending standards, what is and isn't allowed for banks by supervisory authorities in terms of offering loans to consumers is currently clearly specified and understandable for both parties. As a result, in terms of explainability the lending standards are the basis (and thereby the explanation) for rejection of most loans of consumers. As for the edge cases where (within the lending standards) newer AI models might give a different recommendation compared to the traditional models of banks, explainable AI would be especially important to give insight into exactly what causes the deviation from traditional models. Due to the current simplicity of the utilized models, this is at the moment not yet a concern, as also stated by the interviewees. Interviewees at a bank indicated that automated explainability towards consumers (loan applicants) is in principle possible due to the high level of interpretability of the models. Currently, in most cases there is a human-in-the-loop (the advisor) who provides the customer with information and acts as a potential ethical safeguard.

## 4.2 Credit risk management

The AI system of the first bank in the credit risk management use case follows an AIRB (advanced internal rating-based) model for the bank's residential mortgage portfolio (a capital model). It predicts a probability of default for each mortgage customer and a prediction of loss-given-default for each customer. The model uses around 10-15 variables and is based on logistic regression. There is no interaction with any consumer based on the model, it is only used internally. The main stakeholders for explanations are the internal "first line" and the supervisory authority. More advanced AI is expected to potentially be able to lead to better performance, however, the interviewees reported apprehension to use more complex models due to the expected long and time-consuming process to get approval both internally and externally from supervisory authorities.

From the interview with the first SA, it became apparent that regulations such as capital requirements regulations (CRR [28]) heavily determine the boundaries for what type of AI systems can be used in this use case. Predominantly, logistic regression models are used across all financial institutions. Models that are more complex may not meet requirements like traceability and replicability. Another requirement for credit risk models is to demonstrate "experience" in applying a model. In practice, this means that the model must be used as a shadow model for at least three years before approval can be given. Banks are conducting plentiful research and pilots into AI in credit risk, but the regulations are a limiting factor for further implementation. Currently, AI in credit risk does not appear to lead to sufficient benefit compared to the challenge of getting its use approved within the current regulatory framework to make it worthwhile. It was indicated that the bank first to implement a new AI method must assume it takes at least a year and a half before approval is granted.

In summation, in credit risk management strict requirements are heavily embedded in regulations like CRR. Credit risk management forces 'transparent by design' models, therefore, xAI is less of an issue as AI models that are not inherently transparent are simply not used. Regulations/supervisory authorities are slow to change on credit risk, possibly to the more international nature and societal importance of regulation in this use case. Changing these regulations to allow for AI systems that are more complex will be an incremental process that takes time and trust in the safety of such systems.

### 4.3 Anti-money laundering (AML)

For the first bank the use case of anti-money laundering (AML) involved an AI system developed to detect fraudulent activity in corresponding banking transactions. The AI system consists of two algorithms (models): a deduplication algorithm and a classification algorithm. As AML investigators check the flagged transactions, there is a human-in-the-loop. The AML investigator receives explanations (e.g., the most important features leading to a flagging) as part of the outcome of the AI system. The xAI tool SHAP [21] was used with output provided to the investigator. As such, the investigator can be said to be main stakeholder for explanation in this use case. Explanation, in a broader sense, to other stakeholders is done via technical documentation and various internal processes.

The use case of the second bank concerns machine learning (ML) used for transaction monitoring. In the past, transaction monitoring was only done rule-based. Currently, multiple ML models are used in conjunction with a rule-based methodology. For instance, there is a supervised AI model that is used as noise reduction (i.e. reduces false positives) on the output of the rule-based system. Furthermore, there is also a supervised model that gives customers scores based on suspicion of money laundering practices and an unsupervised anomaly detection AI model. The output of the models is intended for transaction monitoring analysts who have expertise in recognizing integrity risks. These analysts are generally not concerned with assessing the quality of model output, which is done by quality assurance analysts. The ML model output includes extensive information (which can be considered explanation) about suspicious situations, e.g., indicating the most relevant features, as opposed to rule-based systems. This explainability aspect of these (modern ML) models is thus an important part of the subsequent analysis done by the analyst. This analyst also uses a multitude of other data (sources) outside the detection models for further verification. The analyst can be seen as the human-in-the-loop in this use case, and as the most important stakeholder in need of explanation. Notably, results of the ML-models are improved over the traditional models: both fewer false positives and fewer false negatives (thus more suspicious transactions are reported).

Interviewees indicated that both internally for banks, but also for supervisory authorities, a change of mindset is required to transition from the traditional way of thinking in thresholds (contained in business rules), to more probabilistic thinking about the features of an AML case (contained in modern ML methods). With the latter, explanations can be more complex, but should not be of less quality.

The first SA, in the case of AML, is tasked with ensuring that banks comply with the Anti-Money Laundering and Anti-Terrorist Financing Act [29]. Currently, this SA does not impose any requirements on what type of AI system is used for AML as long as it can be properly explained both to the supervisory authority and internally. Exactly what sufficient explanation is for which type of AI system is not defined by the SA but assessed on a case-to-case basis, due to the highly varying contexts in which AI is used. For the time being, there is also no framework in which explainability is defined, which is directly applicable to this use case. In the context of controlled business operations, a bank must be able to explain how its systems work. If a bank cannot explain an AI system, both to the supervisory authority and internally, as there may be uncontrolled business operations the bank does not sufficiently manage its risks.

In summary, AML was indicated to be one of the use cases that can benefit most from AI in terms of improving results while also being the use case in which the supervisory authorities allow the most room for the use of novel AI methods. So far, the issue of explainability did not hinder the deployment of more complex AI systems in this use case. The internal AML analyst/investigator is viewed as the most important stakeholder regarding explanations by the banks. This investigator is trained to work with and understand model output, which can be seen as a form of, or bringing about of, explainability.

### 4.4 General

One of the main findings, reported throughout the interviews, is that explainable AI is high on the agenda of banks and supervisory authorities. Within banks, it either is or is planned to be an aspect of an ethical framework used within the organization. Such a framework generally builds on existing principles or procedures (not related to AI specifically), but there is a trend towards more unification of principles and a more explicit focus on AI. For supervisory authorities, explainability is not exclusively an ethical concern, as it is also relevant from a prudential and legal perspective (e.g., a prudential or legal framework such as CRR, lending standards, and the GDPR). As such, explainability is relevant to a wide range of supervisory authorities in the financial sector among which the two in this study, but also including, e.g., data protection supervisory authorities.

The use of complex AI systems by banks is increasing although often still limited, mainly still using basic methods such as logistic regression. The use case of AML is a notable exception where more varied and advanced AI models are used. In the plenary session, the following reasons for the slow adoption of AI were mentioned: 1) The time needed to become familiar with and implement complex models and especially xAI methods (such as SHAP and LIME [21,22]), which have emerged only in the last years. Deciding what xAI method to choose, and how to implement it, is a challenging process as xAI is still developing rapidly and in a short period new methods might make a current xAI method obsolete. 2) Uncertainty as to whether financial regulations (such as lending standards, CRR) or the supervisory authority would allow the use of novel AI. 3) Traditional models are deemed adequate for many use cases. 4) Internal hesitation to implement complex AI systems in customer facing applications. 5) AI systems that are more complex are difficult to maintain and monitor over time.

As for the types of information that can serve as the basis for explanations it could be noted that across all use cases the supervisory authorities indicated they are interested in the full range of types of information, while the interviewees from banks generally indicated only a subset per use case was relevant (see Table 1).

**Table 1.**: Responses of SAs and banks on the importance of the types of information that can potentially underpin an explanation for supervisory authorities per use case. A plus-sign (+) indicates a positive, a minus-sign (-) a negative, and both (+/-) indicates a partial importance. Note that each of the three banks only partook in two use case interviews, and thus two banks responded per use case, except for the credit risk use case where only interviewees of one bank filled in this list.

| | Consumer Credit | | | Credit Risk | | AML | | |
|---|---|---|---|---|---|---|---|---|
| | SAs | Bank | Bank | SAs | Bank | SAs | Bank | Bank |
| The reasons, details, or underlying causes of a particular outcome | + | - | + | + | - | + | - | + |
| The data and features used as input to determine a particular outcome | + | + | + | + | + | + | - | + |
| The data used to train and test the AI system | + | + | + | + | + | + | - | + |
| The performance and accuracy of the AI system | + | - | + | + | - | + | + | + |
| The principles, rules, and guidelines used to design and develop the AI system | + | + | + | + | + | + | + | + |
| The process that was used to design, develop, and test the AI system | + | + | + | + | + | + | + | + |
| The process of how feedback is processed | + | - | + | + | - | + | + | + |
| The process of how explainers are trained | + | + | + | + | + | + | - | + |
| The persons involved in design, development, and implementation of AI system | + | - | + | + | - | + | - | +/- |
| The persons accountable for development and use of the AI system | + | + | + | + | + | + | - | + |

## 5 Discussion and conclusions

The main finding of this study is that there appears to be a disparity between the supervisory authorities (SAs) and the banks regarding the desired scope of explainability required for the use of AI in finance. This is exemplified by responses by these two types of organization on what types of information are required by SAs in the various use cases (visible in Table 1). SAs indicate all types of information are relevant while banks indicate only a subset is relevant. Various laws and regulations already explicitly or implicitly impose requirements on the explainability of information systems, regardless of whether they are AI systems or other classes of systems. However, the use of AI systems brings with it a new type of ethical, social, and legal challenges in addition to the direct technical challenge of opening the black box of non-interpretable models [8,9,23,30]. Therefore, it seems warranted to further explore how this disparity should be addressed.

The financial sector could perhaps benefit from clear differentiation between technical (model) explainability requirements and explainability requirements of business operations, applicable laws, and regulations on this topic of AI. A similar bifurcation as can be made for transparency (process transparency and model transparency [23]) might be useful for the xAI field: for instance, "AI model explainability" and "AI system explainability". The first of these relating to a set of techniques and methods that are directly used to better understand the AI model and how its input relates to its output. The second of these relating to the broader concept of explainability that views the AI model as embedded in a system or a set of systems or processes. Whether a black box houses a deterministic machine learning system, or whether a (larger) black box houses a complex system of processes and various agents interacting with an AI, both require explanation [25]. In the first case the questions will be more like "how does this input lead to this output", the opening of the traditional black box AI. However in the second case questions could be: "how is this process designed?" or "who is responsible for the data quality?".

Most interviewees, especially the technical (i.e. model developers) associated explainable AI with the technical tools that have been developed in the last few years, that focus on explaining the model in a low-level fashion. While technical tools, such as SHAP [21], give additional information about the operation of a model, they do not answer how such information in general is conveyed understandably to a stakeholder, by means of an explanation suited to that stakeholder [9]. Additionally, these tools are often post-hoc or after the fact [13]. Like requirements as privacy, security, and fairness, explainability should require attention from the onset of the design of an information system, "explainability by design" [31,32].

It should be noted that several factors could have made the disparity (seen in Table 1) larger than it is in actuality. Firstly, the interviewees at the bank might not have the same understanding about the laws and regulations as interviewees from the SAs had. Another explanation for the disparity is that it is difficult to translate laws and regulations into precise requirements for information systems and AI systems in particular [33], thus for novel developments very broad ranges of requirements are assumed. The exact reason for the disparity found in this study is certainly a worthwhile topic of future research as well as for subsequent coordination and collaboration between supervisory authorities and regulated entities on topics such as transparency, explainability, and associated definitions.

The requirements regarding explainable AI reported in the interviews varied widely per use case and stakeholder. This limits the possibility of quickly creating a generic framework or checklist for AI in finance that covers all or most bases. Subsequent research could first explore a single use case to create a full picture of the explanation requirements and what information is relevant for which stakeholder given a range of possible AI models. Subsequently, mapping stakeholders to xAI methods [19,21,22] to see how they can be helped can be a valuable avenue of research that can produce practical instruments for the implementation of xAI.

This study has several limitations that should be noted. First, we only interviewed employees of a subset of the Dutch financial sector, three banks and two supervisory authorities. In addition, we only spoke to a total of 21 employees across the five organizations. Furthermore, we only touched the surface in the examination of the use cases with interviews as the main method to collect data. More in-depth studies are necessary to confirm and extend our findings and to determine whether our findings hold across different geographies.

We found banks are hesitant to put complex AI models into practice in their primary business processes for the lending and credit risk use cases. Interestingly, supervisory authorities indicated that they in principle do not restrict the use of specific types of AI systems. However, laws and regulations such as lending standards and CRR impose explainability requirements which limit the choice of AI methods beforehand. This might be a chicken-or-the-egg type problem, in which banks are unclear what regulators would precisely allow and therefore do not develop a certain AI solution (based on a certain model), while regulators wait for banks to put AI systems into practice before they can clearly say which type of model is allowed and which is not. To counteract this, in the plenary session it was proposed to increase communication between banks and SAs, also in the development process of new AI models.

Notably, in the consumer credit and AML use cases, the use of novel AI methods went hand in hand with the ability to leverage more (types) of data in addition to the ability to use historical data. This is a clear advantage of these novel AI methods over the traditional business rule systems and might explain the increased performance that was reported in these use cases.

The application of AI at banks for the three use cases is currently only focused on internal stakeholders, such as the investigators in the AML use case or the mid-office employees in the consumer credit use case. The fact that there is a human-in-the-loop was reported as a positive, as this offered an additional safeguard before action was taken based on the AI output. In the future, more familiarity with (fully) automated AI systems might lead to banks deploy more customer-oriented AI.

This is one of the first studies that provides practical insight in the application of xAI in the context of use cases and AI systems in the financial sector. It demonstrates that a wide range of aspects requires attention when designing and building AI systems, and that explainability cannot be considered as a merely technical challenge nor a one-size-fits-all solution. For financial law and policy makers, this research illustrates that financial laws and regulations have an impact on the design of information systems and in particular, AI systems.

In conclusion, there appears to be a disparity between the perspectives as provided by the interviewees of the banks and those of the supervisory authorities for the use cases investigated in this study. Namely, the supervisory authorities view explainability of AI in a wider fashion. Potentially, this can be reframed as the supervisory authorities requiring explanation of the AI model as embedded in a broader system, explicitly or implicitly part of financial laws and regulations. On the other hand, the regulated entities (i.e. the banks in this study) tended to view explainable AI more as a requirement of only the AI model. A clear differentiation between technical AI (model) explainability requirements and explainability requirements of the wider AI system in relation to applicable laws and regulations can potentially be of benefit to the financial sector and help in the communication between supervisory authorities and banks.